%% file: main.tex
\pdfoutput=1

\documentclass[11pt]{article}

\usepackage[final]{ACL2023}

\usepackage{times}
\usepackage{latexsym}
\usepackage{enumitem}

\usepackage[T1]{fontenc}

\usepackage[utf8]{inputenc}

\usepackage{microtype}

\usepackage{inconsolata}

\usepackage{url}
\usepackage{times}
\usepackage{latexsym}
\usepackage{graphicx}
\usepackage{todonotes}
\usepackage{subcaption}
\usepackage{amsmath}
\usepackage{booktabs}
\usepackage{multirow}
\usepackage{todonotes}
\usepackage{wasysym}
\usepackage{tabularx}
\usepackage{pgfplots}
\usepackage{amssymb}
\usepackage{pifont}

\newcommand{\SQuArE}{SQuArE}

\usepackage{enumitem}
\setlist[itemize]{align=parleft,left=0pt..1em}

\newcommand{\mypara}[1]{\noindent\textbf{#1}}

\title{SQUARE: Automatic Question Answering Evaluation using \\ Multiple Positive and Negative References}

\author{Matteo Gabburo$^{1}$\thanks{\ \ Work done as an intern at Amazon Alexa AI}\ , Siddhant Garg$^{2}$,  \textbf{Rik Koncel Kedziorski$^{3}$\thanks{\ \ Work completed at Amazon Alexa AI}\ , Alessandro Moschitti$^{2}$}\\
$^{1}$University of Trento , $^{2}$Amazon Alexa AI, $^{3}$Kensho Technologies, Inc. \\
\texttt{matteo.gabburo@unitn.it} \\ \texttt{\{sidgarg,amosch\}@amazon.com} \\
\texttt{rikka@kensho.com} \\
}

\begin{document}
\maketitle
\begin{abstract}
Evaluation of QA systems is very challenging and expensive, with the most reliable approach being human annotations of correctness of answers for questions. 
Recent works (AVA, BEM) have shown that transformer LM encoder based similarity metrics transfer well for QA evaluation, but they are limited by the usage of a single correct reference answer. We propose a new evaluation metric: {\SQuArE} (Sentence-level QUestion AnsweRing Evaluation), using multiple reference answers (combining multiple correct and incorrect references) for sentence-form QA. We evaluate SQuArE on both sentence-level extractive (Answer Selection) and generative (GenQA) QA systems, across multiple academic and industrial datasets, and show that it outperforms previous baselines and obtains the highest correlation with human annotations.
\end{abstract}


\section{Introduction}
\label{sec:introduction}

Automatic evaluation of Question Answering systems to gauge correctness of an answer for a question is a challenging task. This task is important for preserving a quick velocity in evaluating and development of new QA systems, and creating large high quality training corpora for LLM-based QA systems. The most common approach for this task is to obtain human annotations of correctness of answers for questions, which is slow, expensive, and challenging (annotating complete answer sentences for questions has been shown to achieve poor inter-annotator agreement).

Span extraction (MR) based QA systems are typically evaluated using token matching metrics such as EM (Exact Match) or F1, however, these cannot be extended for evaluating complete sentence-form answers coming from Answer Sentence Selection (AS2) systems~\cite{garg2020tanda,di-liello-etal-2022-pre,di-liello-etal-2023-context}. Token/segment-level similarity metrics such as EM, F1, BLEU, etc. fail to capture the semantic coherence between entities/concepts of the answer sentence and the question. Recently, AVA~\cite{vu_ava_2021} and BEM~\cite{bulian2022tomayto} have proposed transformer LM encoder based similarity metrics for sentence-form extractive QA evaluation by encoding the question, target answer (which needs to be evaluated) and a reference answer (which is treated as a gold standard (GS)).

\begin{figure}[t]
\includegraphics[width=\linewidth]{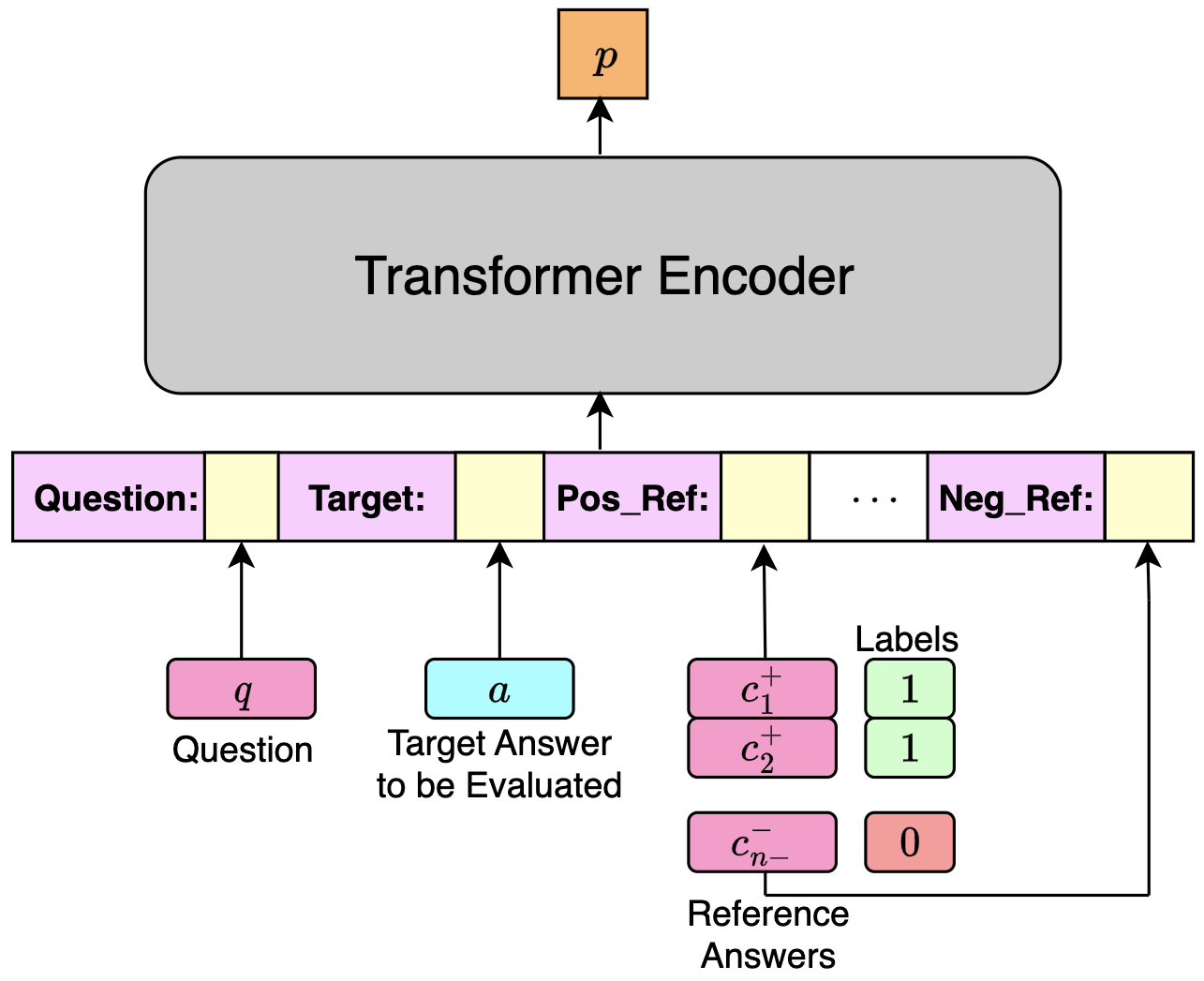}
\vspace{-0.5em}
\centering
\caption{ An illustration of {\SQuArE}: an automatic question answering evaluation metric that uses multiple references: positive and negative to evaluate the correctness of a target answer for a particular question.}
\label{fig:ava}
\vspace{-0.5em}
\end{figure}

One of the major limitations of AVA/BEM is the use of a single reference answer. There are several types of questions that have multiple diverse correct answers, other questions that have relevant information spread across multiple reference answers, and other ambiguous/under-specified or opinion seeking questions that may have several possible answers (we motivate this with examples in Section~\ref{sec:methodology}). Additionally, AVA/BEM only use information from a correct reference answer for evaluating a target answer, but information and semantics from an incorrect reference answer (which are readily available for several datasets) can also help refine the accuracy of the prediction.

Motivated by the above shortcomings of AVA/BEM, we propose {\SQuArE} (Sentence-level QUestion AnsweRing Evaluation), a supervised transformer LM encoder based automatic QA evaluation metric that uses multiple reference answers by combining multiple correct and incorrect answers to assign a correctness score for an answer to a question. We evaluate {\SQuArE} on four sentence-level extractive QA datasets, and show that it outperforms previous baselines and achieves the highest correlation with human annotations.

The last few years have seen several research works~\cite{hsu2021answer,DBLP:journals/corr/abs-2110-07150,gabburo2022} transition from extractive sentence-form QA towards generating natural sounding sentence-form answers. This paradigm (termed GenQA) synthesizes information using different pieces of information spread across many relevant candidates (while suppressing any irrelevant information) to improve the answering accuracy and style suitability.  AVA/BEM have only been evaluated on extractive QA, and not for GenQA, so it is unclear if a transformer encoder based semantic matching metric will correlate with human annotations on a sentence-form generated answer. We strengthen the generality of {\SQuArE} as a QA evaluation metric by showing that it outperforms the AVA/BEM baselines for GenQA systems in addition to extractive QA systems. We will release the code and trained model checkpoints for {\SQuArE} at \url{https://github.com/amazon-science/square} for the NLP and QA community to use our automatic QA evaluation metric.

\section{Related Work}
\label{sec:relatedwork}

\mypara{Automatic Text Similarity Evaluation:} Token/N-grams level similarity metrics like BLEU~\cite{papineni_bleu_2001} and ROUGE~\cite{lin-2004-rouge} are not suitable for QA evaluation, and have been shown to achieve poor correlation with human judgements~\cite{reiter_structured_2018,gabburo2022}. \citet{kusner2015wmd} propose using a distance function between word embeddings for text similarity. Other research works~\cite{kusner2015wmd,clark-etal-2019-sentence} have proposed evaluation metrics based on Wasserstein distance. Recent years have seen a number of automatic evaluation metrics being proposed for Neural Machine Translation (MNT) and summarization tasks like BERT-Score \cite{zhang_bertscore_2020}, BLEURT \cite{sellam_bleurt_2020}, COMET \cite{rei_comet_2020}, etc. that use contextual embeddings from transformer encoders. Similar approaches extend for text style~\cite{wegmann_does_2021} and summarization~\cite{cao_factual_2020, zeng_gradient-based_2021}.

\mypara{QA Evaluation:} For entity level span-extraction MR tasks, \citet{yang-etal-2018-adaptations} adapt BLEU, ROUGE for answer comparison, with a focus on “yes-no” and “entity” questions. \citet{si-etal-2021-whats} mine entities from KBs to use them as additional gold answers for MR tasks, our approach shares this intuition of using multiple diverse reference answers for evaluation. \citet{chen-etal-2019-evaluating} propose a modification of BERTScore for QA by using the question and the paragraph context along with the answer. Empirically however, they demonstrate that for extractive MR tasks, F1 works as a reasonable metric, but this does not transfer well for generative QA. \cite{pmlr-v133-min21a} uses human annotations to evaluate correct answers that are not contained in the GS answer. For sentence-level extractive QA (AS2), AVA \cite{vu_ava_2021} and BEM \cite{bulian2022tomayto} are two recently proposed learned metrics. 

\section{Methodology}
\label{sec:methodology}
Being a knowledge-intensive task, automatic QA evaluation typically requires leveraging knowledge from external sources to evaluate correctness of answer (e.g., Knowledge Bases, Gold standard reference answers). We can formalize automatic QA evaluation with the notation: \(f(q, a, c){\xrightarrow{}}p\) , where \textit{f} is the automatic evaluation function applied to question \textit{q}, target answer \textit{a} and reference context \textit{c}, and outputs a correctness score \(\textit{p} \in [0,1]\). 

Previous works (AVA, BEM) show that using a single GS reference answer as the context \textit{c} achieves higher correlation with human annotations that only using $q$ and $a$. In this paper, we propose a supervised learned metric {\SQuArE} that enriches the reference context \textit{c} for QA evaluation using: (i) multiple gold standard references, and (ii) negatively annotated answers as negative references.

\mypara{Multiple Reference Answers} In AVA/ BEM, using a single correct reference limits the evaluation scope of QA system predictions. 

\begin{itemize}[noitemsep,nolistsep]
    \item Several types of questions may have multiple and diverse correct answers: for example \textit{"What is a band?"} is correctly answered by both \textit{"A flat, thin strip or loop of material, used as a fastener"} and \textit{"A band is a group of people who perform instrumental and/or vocal music"}
    \item Knowledge seeking questions may have pieces of relevant information spread across multiple references: for example \textit{"Who is Barack Obama"} can be answered by combining information across multiple answers \textit{"He served as the 44th president of the U.S. from 2009-2017"}, \textit{"He was a member of the Democratic Party, and served as a U.S. senator from 2005-2008"}, etc. 
    \item For ambiguous/under-specified questions that do not have a single correct answer or opinion seeking questions, using a single GS reference answer can be limiting and provide an incorrect evaluation of the answering capability of a QA system. Consider the question \textit{"When is the next world cup"} for which both the answers \textit{"The next FIFA football world cup is in 2026"} and \textit{"The next ICC cricket world cup is in 2023 in India"} are correct as the questions fails to specify the name of the sport (many more possible answers).
\end{itemize}


\mypara{Negative Reference Answers} An automatic QA evaluation system can use the information and semantics from an incorrect answer to help refine the accuracy of its prediction. Consider the question \textit{"Which movies of Dwayne Johnson released in 2017"} with the positive reference \textit{"Dwayne `The Rock' Johnson starrer Baywatch premiered in 2017"}. Only using this reference, both the answers \textit{"Baywatch and Jungle Cruise"} and \textit{"The Fate of the Furious and Baywatch"} appear to be equally correct for this question. However when we add in an incorrect reference for the question \textit{"Jungle Cruise is a movie starring the Rock and Emily Blunt that released in 2021"}, the automatic QA evaluation can identify that the second answer is probably more correct than the first one. Several sentence-form extractive QA datasets such as ASNQ~\cite{garg2020tanda}, WikiQA, TREC-QA, etc. have a large number of negatively labeled answer candidates for each question, which can be exploited for automatic evaluation of QA systems for these datasets. 

\mypara{{\SQuArE}} Motivated by the above reasons, we modify the context $c$ of automatic evaluation \(f(q, a, c){\rightarrow}p\) to include a combination of $n_{+}$ correct and $n_{-}$ incorrect reference answers, i.e, \(c \ {:} \ {c^{+}{=}\{c_{1}^{+}, ...,c_{n_+}^{+}\} \cup c^{-}{=}\{c_{1}^{-}, ...,c_{n_-}^{-}\}}\). During supervised learning, {\SQuArE} learns to minimize the semantic distance between a correct target answer from the set of correct references $c^{+}$ and maximizing the semantic distance from the set of incorrect references $c^{-}$.
We prefix a prompt (\textit{Pos\_Ref} / \textit{Neg\_Ref}) to each reference to indicate the correctness/incorrectness of the reference to the model. Specifically, a $(q, a, c^{+}, c^{-})$ input for {\SQuArE} is encoded as \textbf{"}\textcolor{blue}{Question:} $q$ \textcolor{blue}{Target:} $a$ \textcolor{blue}{Pos\_Ref:} $c_{1}^{+}$ ${\cdots}$ \textcolor{blue}{Pos\_Ref:} $c_{n_{+}}^{+}$ \textcolor{blue}{Neg\_Ref:} $c_{1}^{-}$ ${\cdots}$ \textcolor{blue}{Neg\_Ref:} $c_{n_{-}}^{-}$\textbf{"} as illustrated in Figure~\ref{fig:ava}. 

The choice of reference answers can create biases in automatic QA evaluation. For a given question, collecting a set of diverse reference answers and ensuring they exhaustively cover all the concepts needed to answer the question is challenging and very expensive. In this paper, we utilize existing annotated answer candidates (both positive and negative) in high-quality labeled datasets as references. Extending automatic QA evaluation to previously unseen questions (without any references) is a challenging open problem in NLP QA.

\section{Experiments and Results}
\label{sec:experiments}

\subsection{Datasets}
\label{sec:data}


\mypara{WQA} Web Question Answers (WQA) is a public dataset \cite{zhang-etal-2021-joint} containing 149,513 questions, each associated with ${\sim}$15 answer candidates retrieved from a large-scale web index with human annotations.

\mypara{WikiQA} A small AS2 dataset~\cite{yang2015wikiqa} with questions from Bing search, and answers extracted from Wikipedia. We use the most popular clean setting (questions having at least one positive and one negative answer).

\mypara{TREC-QA} A small AS2 dataset~\cite{wang2007jeopardy} containing factoid questions. We only retain questions with at least one positive and one negative answer in the development and test sets.

\mypara{IQAD}  A large scale Industrial QA Dataset containing non-representative de-identified user questions from a commercial personal assistant. IQAD contains 10k questions, and $\sim$200 answer candidates retrieved for each question using a large scale web index that contains over 100M web documents. Results on IQAD are presented relative to a baseline to comply with company policies.

\mypara{GenQA-MTURK} This dataset is composed of 3k questions from 3 datasets (1k each): MS-MARCO \cite{bajaj2018ms}, WikiQA and TREC-QA using GenQA models evaluated in \cite{hsu2021answer,gabburo2022}. For each question we generate an answer using 8 different GenQA models (details in Appendix~\ref{apx:datasets}) based on T5-Large. We annotate all the answers of this dataset for their correctness, using MTurk using $5$ independent annotations for each QA pair. We use majority voting over the 5 annotations for each QA pair.

\mypara{Answer Equivalence (AE):} A question answering dataset released by \citet{bulian2022tomayto}  where each sample contains a question, a candidate answer (typically short answers), and a positive reference (typically entity-based) carefully selected to avoid the candidate-reference exact match (EM).

\input{Tables/big_table}

\subsection{Models and Baselines}
\label{ssec:baselines}

We use DeBERTaV3-Large \cite{he2021debertav3} for {\SQuArE}, and compare with three baselines (proposed in AVA/BEM): \textbf{QT: Question-Target} that takes input a question and the target answer, \textbf{TR: Target-Reference} that takes input a reference GS answer and the target answer, and \textbf{TQR: Target-Question-Reference} that takes as input a question, the target answer and a reference GS answer. For our experiments, we set the total number of reference $(n_{+}){+}(n_{-}){=}5$ per question.

We also compare {\SQuArE} against two additional baselines: (i) \textbf{BEM}~\cite{bulian2022tomayto}, a recently released reference-based automatic evaluation metric (trained on the AE dataset), and (ii) a large language model (\textbf{LLM}) based approach using two versions of the Falcon~\cite{falcon40b} model. For fair comparison with the baselines, we perform evaluation in the zero-shot setting for the WikiQA and TrecQA datasets, and after fine-tuning on the AE dataset. For more details on the implementation of these baselines, refer to Appendix~\ref{apx:bemandllm}.

\subsection{Results}
\label{ssec:results}

We present results comparing {\SQuArE} with the baselines on large datasets (from both extractive QA: AS2 and generative QA: GenQA) in Table~\ref{tab:publicres}. Using GS human annotations for each dataset, we compute accuracy, Area Under the Curve (AUROC), and Pearson Correlation of each automatic QA metric. We observe that on all datasets, {\SQuArE} significantly outperforms the baselines and achieves the highest accuracy and AUROC with human annotations.  

\mypara{Zero-shot Setting:} To show strong generalization to out-of-distribution datasets (zero-shot setting), we train {\SQuArE} and the other baselines on the WQA dataset, and use this for evaluation on other datasets. Specifically, we use two small datasets: WikiQA and TREC-QA (exploring both extractive: AS2 and generative settings), and one large dataset MS-MARCO. Results presented in Table~\ref{tab:publicres2} highlight that {\SQuArE} achieves the highest accuracy and correlation with human annotations.

\mypara{Comparison with BEM and LLMs:} We present comparison with BEM and LLM baselines in Table~\ref{tab:comparisonwithbem} on WikiQA, TrecQA and Answer Equivalence (AE) datasets. On the WikiQA and TrecQA datasets, the results show that {\SQuArE} outperforms both the baselines, which stems from (i) the usage of multiple references, and (ii) the references for these datasets being complete sentences in comparison to entities/short-answers which are used for training BEM. On the AE dataset, zero-shot {\SQuArE} (which is trained on the WQA dataset) performs inferior (0.572 vs 0.897 in accuracy) to the BEM baseline (which is trained on the AE dataset). This drop in zero-shot performance of {\SQuArE} compared to BEM can be attributed to (i) the lack of multiple references, and (ii) the references in AE being of a different style/format than those used for training {\SQuArE} (entities/short answers v/s complete sentences). On fair comparison (when {\SQuArE(AE)} is fine-tuned on the AE dataset), it is able to beat the BEM baseline in both accuracy (0.908 vs 0.897) and AUROC (0.966 vs 0.859).

\input{Tables/bemandllm_table}



\begin{table}[h]
\small
\centering
\resizebox{0.95\linewidth}{!}{
\begin{tabular}{lccc}
\toprule
\textbf{Dataset} & \textbf{SQuArE} & \textbf{BLEURT} & \textbf{BERTScore}  \\
\midrule
MS-MARCO & \textbf{0.238} & 0.142 & 0.168 \\
WikiQA   & \textbf{0.425} & 0.219 & 0.233 \\
TrecQA   & \textbf{0.862} & 0.341 & 0.646 \\
\bottomrule
\end{tabular}
}
\vspace{-0.5em}
\caption{Pearson Correlation of evaluation metrics with human annotations on GenQA-MTURK.}
\label{tab:correlation_ava_automaticmetrics}
\end{table}

\mypara{Comparison with text similarity metrics:} We also compare {\SQuArE} with learned text similarity metrics: BLEURT and BERTScore in Table~\ref{tab:correlation_ava_automaticmetrics}. Results show that SQuARe achieves a higher correlation with manual annotations than BLEURT and BERTScore. For complete details, see Appendix~\ref{apx:correlation_similarity_metrics}.

\vspace{-0.3em}
\subsection{Ablation studies}
\label{sec:ablation_studies}
\vspace{-0.2em}

To assess the improvements from different design choices used in {\SQuArE}, we conduct ablation studies to show how the use of negative and multiple references improves the performance and correlation with human annotations. To perform these studies we pick one dataset (WQA) and present comparisons in Tab.~\ref{tab:ablations}.

\input{Tables/ablations_table}

\mypara{Usage of Negative references:} To support our claim that using negative references can improve the automatic QA evaluation, we compare two additional models/baselines: (i) AVA-TQR(-) which refers to an AVA baseline which only uses a single negative reference, and (ii) {\SQuArE}(+) which refers to a {\SQuArE} model which only uses multiple positive references. On comparison with results in Table~\ref{tab:publicres}, AVA-TQR(-) outperforms both AVA-QT (model without references) and AVA-TR (model without the question). This validates our intuition on the importance of negative references. {\SQuArE}(+) outperforms the AVA-TQR baseline, but performs inferior to the {\SQuArE} using a combination of both positive and negative references, thereby validating our claim that the combination of positive and negative references improves the accuracy and the generalizability of {\SQuArE}. 

\mypara{Number of references:} We hypothesize that higher number of labeled references help with improved correlation of {\SQuArE} with human evaluation. To support this intuition, we present an ablation study where we vary the total number of references from 5 per question to: (i) using 3 references per question, and (ii) randomly sampling ${\in}[1,5]$ references per question. We observe that {\SQuArE} using 5 references outperforms {\SQuArE} using 3 references (0.833 v/s 0.821 in accuracy), while {\SQuArE} using a random sample of ${\in}[1,5]$ references (0.820 accuracy) performed comparable to {\SQuArE} using 3 references.

\section{Conclusion}
\label{sec:conclusion}

In this paper, we propose {\SQuArE} transformer LM encoder-based learned metric that uses multiple reference answers (positive + negative) for automatically evaluating sentence-level QA systems. We evaluate sentence-level extractive QA: AS2 and answer generation (GenQA) systems across multiple academic and industrial datasets and show that {\SQuArE} achieves the highest correlation with human annotations beating previous baselines.

\section*{Limitations}
\label{sec:limitations}

Our approach of training QA evaluation metrics requires access to large GPU resources for training large transformer encoders such as DeBERTa, etc. For the experiments in this paper, we only consider datasets from the English language, however we conjecture that our techniques should work similarly for languages with a similar morphology. Since {\SQuArE} is a learned evaluation metric based on large transformers, it might be challenging to effectively learn in a scarce-data setting. While we have shown impressive zero-shot evaluation results in Table~\ref{tab:publicres2}, extending to a completely new data domain/new language might be challenging for {\SQuArE} to adapt to without access to any labeled data. As visible from Tables~\ref{tab:publicres} and \ref{tab:publicres2}, {\SQuArE}'s accuracy on human annotations is in the range of 80-90\%, highlighting that there is still a gap with respect to human evaluation. For safety critical applications, human evaluation still remains the best means to evaluate Question Answering systems.

\bibliography{anthology,custom}
\bibliographystyle{acl_natbib}

\clearpage
\appendix

\section*{Appendix}

\section{Experiment Details}
\label{apx:experimentalsetting}

In this section we describe the parameters used to train and reproduce our models, and the computational environment we used to compute the results.

\subsection{{\SQuArE} Training}
\label{apx:squaretraining}

We train {\SQuArE} train our models starting from DeBERTaV3-Large model \cite{he2021debertav3}, similar to \cite{gabburo-etal-2023-learning}. We experimented with different parameters, and we found the best combination of parameters training the model for $20$ epochs on every dataset using a batch size of $32$, $fp32$ and Adam \cite{Kingma2015AdamAM} as optimizer with a learning rate of $1e-06$. At the beginning of each epoch we shuffle the train set. We select the best checkpoint by selecting the best AUROC (Area Under the Curve) on the validation set. We train the QT, TQR, and TR using the parameters described in \cite{vu_ava_2021}.

\subsection{BEM and LLM Baselines}
\label{apx:bemandllm}
For BEM, we use the evaluation script and checkpoints from the original paper\footnote{\url{https://github.com/google-research-datasets/answer-equivalence-dataset}}, while for the LLM-based \textit{Falcon} experiments, we consider the originally released checkpoints\footnote{\url{https://huggingface.co/tiiuae/falcon-7b-instruct}}\footnote{\url{https://huggingface.co/tiiuae/falcon-40b-instruct}}. For both the approaches (BEM and LLM) we use the \textit{AVA-TQR} data setting, by providing the question and a positive reference, and asking the model to evaluate the correctness of the target answer. To fine-tune {\SQuArE} on the AE dataset, we use the same parameters as for WQA, described in Section~\ref{apx:squaretraining}.

\subsection{Computational Environment}
\label{apx:computationalenv}
We trained our models using 8 Nvidia V100 with $32$Gb each. Our code is written in PyTorch \cite{pytorch} using Hugging Face \cite{lhoest-etal-2021-datasets, wolf-etal-2020-transformers}. We compute results using metrics developed in Scipy \cite{scipy}.

\section{GenQA-MTURK details}
\label{apx:datasets}

We designed this dataset starting from a reduced sample of the MS-MARCO NLG development set, the test set of WikiQA and the TrecQA test set proposed from \cite{trecqazeyu}. Specifically, we trained 8 different T5-Large models (on MS-MARCO) using training techniques from \cite{hsu2021answer} and \cite{gabburo2022}: (i) Supervised GenQA, (ii) Weak Supervision (WS), (iii) WS+Loss Weighting (LW), (iv) WS+Score-conditioned Input (SCI), (v) WS+Score-conditioned Output (SCO), (vi) WS+SCI+SCO, (vii) WS+LW+SCI+SCO, (viii) Supervised GenQA+WS+LW+SCI+SCO.

\section{Correlation with Similarity Metrics}
\label{apx:correlation_similarity_metrics}

In Section~\ref{sec:experiments}, we study the correlation of {\SQuArE} with other learned metrics as BLEURT and BERTScore. Table~\ref{tab:correlationmetrics_generative_systems} contains results for each metric broken down based on different GenQA models trained using techniques from \cite{hsu2021answer, gabburo2022}. The results show that despite the good performance of the BLEURT and BERTScore metrics, {\SQuArE} has a better correlation with human evaluation. In addition, since our approach can use negative references, it can obtain higher performance than similarity metrics for datasets having a scarce number of positive references.

\begin{table}[h]
\small
\centering
\resizebox{\linewidth}{!}{
\begin{tabular}{@{}lcccccc@{}}
\toprule
\textbf{Dataset} &
  \textbf{Sys} &
  \textbf{Accuracy} &
  \textbf{SQuArE} &
  \textbf{BLEURT} &
  \textbf{BERTScore}  \\ \midrule
\multirow{6}{*}{\textbf{MS-MARCO}}     & 1 & 0.946   & 0.881     & 0.772 & 0.732  \\
                                       & 2 & 0.901   & 0.913     & 0.578 & 0.486 \\
                                       & 3 & 0.958   & 0.779     & 0.569 & 0.485 \\
                                       & 4 & 0.878   & 0.781     & 0.578 & 0.496 \\
                                       & 5 & 0.880   & 0.761     & 0.689 & 0.620 \\
                                       & 6 & 0.906   & 0.768     & 0.714 & 0.658 \\
\midrule
\multirow{6}{*}{\textbf{WikiQA}}       & 1 & 0.804 & 0.513 & 0.514 & 0.406  \\
                              & 2 & 0.803 & 0.316 & 0.447 & 0.331 \\
                              & 3 & 0.826 & 0.323 & 0.491 & 0.360  \\
                              & 4 & 0.791 & 0.306 & 0.480  & 0.350 \\
                              & 5 & 0.804 & 0.557 & 0.669 & 0.623  \\
                              & 6 & 0.837 & 0.609 & 0.579 & 0.510  \\
\midrule
\multirow{6}{*}{\textbf{TrecQA}}       & 1 & 0.874 & 0.769  & 0.563 & 0.324  \\
                              & 2 & 0.976 & 0.830  & 0.557 & 0.409  \\
                              & 3 & 0.968 & 0.932  & 0.482 & 0.498  \\
                              & 4 & 0.871 & 0.643  & 0.491 & 0.420  \\
                              & 5 & 0.968 & 0.868  & 0.628 & 0.442  \\
                              & 6 & 0.976 & 0.917  & 0.580 & 0.595 \\
\bottomrule
\end{tabular}
}
\caption{System-wise evaluation done on three datasets (MS-MARCO, WikiQA, TrecQA) using 6 different systems based on generative question answering models trained on MS-MARCO. The accuracy column is computed using the manual annotations done by professional annotators, while the three remaining columns are the results computed using {\SQuArE}, BLEURT and BERTScore.}
\label{tab:correlationmetrics_generative_systems}
\end{table}


\end{document}

%% file: Tables/big_table.tex
\begin{table}[t]
\centering
\resizebox{\linewidth}{!}{
\begin{tabular}{@{}llcccc@{}}
\toprule
\textbf{Dataset} & \textbf{Technique} & \textbf{\# Refs} & \textbf{Accuracy} & \textbf{AUROC} & \textbf{Correlation} \\ \midrule
\multicolumn{6}{c}{\textbf{Answer Sentence Selection (AS2)}}\\
\midrule
\multirow{4}{*}{WQA}    & AVA-TR  & 1 & 0.734 & 0.809 & 0.716 \\
                        & AVA-QT  & 0 & 0.790 & 0.851 & 0.750 \\
                        & AVA-TQR & 1 & 0.809 & 0.873 & 0.771 \\
                        & SQuArE  & 5 & \textbf{0.833} & \textbf{0.896 } & \textbf{0.793} \\
\midrule
\multirow{4}{*}{IQAD}   & AVA-TR  & 1 & Baseline & Baseline & Baseline \\
                        & AVA-QT  & 0 & +1.94\%  & -0.393\% & +0.682\% \\
                        & AVA-TQR & 1 & +8.02\%  & +5.7\%   & +6.178\% \\
                        & SQuArE  & 5 & \textbf{+22.24\%} & \textbf{+14.01\%} & \textbf{+16.062\%} \\
                        \midrule
\multicolumn{6}{c}{\textbf{Answer Generation (GenQA)}}\\
\midrule
\multirow{4}{*}{MS-MARCO}  & AVA-TR  & 1 & 0.882 & 0.768 & 0.610 \\                                                                                        & AVA-QT  & 0 & 0.882 & 0.777 & 0.623 \\
                           & AVA-TQR & 1 & 0.878 & 0.790 & \textbf{0.636 } \\
                           & SQuArE  & 5 & \textbf{0.895} & \textbf{0.832} & 0.629 \\
\bottomrule 
\end{tabular}}
\vspace{-0.5em}
\caption{Results on WQA, IQAD, MS-MARCO measured using Accuracy, Area under the curve and Pearson Correlation with gold labels. Results on IQAD are relative to AVA-TR baseline (due to data being internal). \# Refs refers to the total number of reference answers used for the metric.}
\label{tab:publicres}
\vspace{-0.2em}
\end{table}

\begin{table}[t]
\centering
\resizebox{0.95\linewidth}{!}{
\begin{tabular}{@{}llccc@{}}
\toprule
 \textbf{Dataset}                & \textbf{Technique} & \textbf{Accuracy}    & \textbf{AUROC}       & \textbf{Correlation} \\ \midrule
\multicolumn{5}{c}{\textbf{Answer Sentence Selection (AS2)}}\\
\midrule
  \multirow{4}{*}{WikiQA } & AVA-TR                                          &  0.701               & 0.633                     & 0.532                           \\
                                                                & AVA-QT     &  0.900               & 0.804                     & 0.637                           \\
                                                                & AVA-TQR    &  0.903               & 0.805                     & 0.632                           \\
                                                                & SQuArE     &  \textbf{0.919}               & \textbf{0.851}                    & \textbf{0.676}                           \\
 \cmidrule(l){1-5}
                               \multirow{4}{*}{TrecQA } & AVA-TR             &  0.911               & 0.913                     & 0.816                          \\
                                                                & AVA-QT     &  0.885               & 0.927                     & 0.737                            \\
                                                                & AVA-TQR    &  0.906               & \textbf{0.972}                     & 0.797                           \\
                                                                & SQuArE     &  \textbf{0.924}               & 0.969                     & \textbf{0.842}                           \\

\midrule
\multicolumn{5}{c}{\textbf{Answer Generation (GenQA)}}\\
\midrule
 \multirow{4}{*}{MS-MARCO} & AVA-TR     &  0.843                                       &    0.683                 &    0.587         \\
                                                                          & AVA-QT     &  0.772                     &    0.693                  &    0.580         \\
                                                                          & AVA-TQR    &  0.839                     &    0.738                  &    0.601         \\
                                                                          & SQuArE &  \textbf{0.845}                         &    \textbf{0.773}                  &    \textbf{0.620}         \\
 \cmidrule(l){1-5}
                                \multirow{4}{*}{WikiQA }    & AVA-TR      &  0.692                     &    0.670                  &    0.602         \\
                                                                          & AVA-QT     &  0.627                     &    0.798                  &    0.667         \\
                                                                          & AVA-TQR    &  0.671                     &    0.811                  &    0.678         \\
                                                                          & SQuArE &  \textbf{0.694}                         &    \textbf{0.819}                  &    \textbf{0.690}         \\
 \cmidrule(l){1-5}
                                \multirow{4}{*}{TrecQA}    & AVA-TR       &  0.847                     &    0.784                  &    0.615         \\
                                                                          & AVA-QT     &  0.709                     &    0.816                  &    0.612         \\
                                                                          & AVA-TQR    &  0.779                     &    \textbf{0.857}                  &    0.647         \\
                                                                          & SQuArE &  \textbf{0.890}                         &    0.818                  &    \textbf{0.671}           \\

\bottomrule 
\end{tabular}}
\vspace{-0.5em}
\caption{Zero-shot evaluation using QA evaluation models trained on WQA. Same metrics used as Table~\ref{tab:publicres}.}
\label{tab:publicres2}
\vspace{-0.2em}
\end{table}

%% file: Tables/bemandllm_table.tex
\begin{table}[t]
\centering
\resizebox{0.95\linewidth}{!}{
\begin{tabular}{@{}llccc@{}}
\toprule
 \textbf{Dataset} & \textbf{Approach} & \textbf{\# Refs} & \textbf{Accuracy} & \textbf{AUROC} \\ \midrule
 \multirow{4}{*}{WikiQA}
 & BEM        & 1 & 0.863  & 0.553 \\
 & Falcon-7B  & 1 & 0.081 & 0.448 \\
 & Falcon-40B & 1 & \textbf{0.963}  & 0.499  \\
 & SQuArE     & 5 & 0.919  & \textbf{0.851}  \\
 \cmidrule(l){1-5}
 \multirow{4}{*}{TrecQA }
 & BEM        & 1 & 0.866 & 0.819  \\
 & Falcon-7B  & 1 & 0.601 & 0.529  \\
 & Falcon-40B & 1 & 0.848 & 0.509  \\
 & SQuArE     & 5 & \textbf{0.924} & \textbf{0.969} \\
 \cmidrule(l){1-5}
 \multirow{3}{*}{AE}
 & BEM         & 1 & 0.897 & 0.959  \\
 & SQuArE      & 1 & 0.572 & 0.718  \\
 & SQuArE(AE)  & 1 & \textbf{0.908} & \textbf{0.966} \\
\bottomrule 
\end{tabular}}
\vspace{-0.5em}
\caption{Comparing SQuArE against BEM and LLM baselines on the WikiQA, TrecQA and AE datasets. The BEM baseline is trained on the AE dataset. We use the same metrics as Table~\ref{tab:publicres}.}
\label{tab:comparisonwithbem}
\end{table}

%% file: Tables/ablations_table.tex
\begin{table}[h]
\centering
\resizebox{\linewidth}{!}{
\begin{tabular}{@{}lcccc@{}}
\toprule
\textbf{Technique} & \textbf{\# Refs} & \textbf{Accuracy} & \textbf{AUROC} & \textbf{Correlation} \\ 
\midrule
 AVA-TQR(-) & 1 & 0.800 & 0.864 & 0.763 \\
 SQuArE(+) & 5 & 0.815 & 0.885 & 0.783 \\

SQuArE & 3 & 0.821 & 0.889 & 0.787 \\
SQuArE & [1,5] & 0.820 & 0.889 & 0.786 \\
SQuArE  & 5 & \textbf{0.833} & \textbf{0.896 } & \textbf{0.793} \\
\bottomrule 
\end{tabular}}
\vspace{-0.5em}
\caption{Ablation studies evaluating the benefits of using negative references, and the impact of number of references on the performance of {\SQuArE}. AVA-TQR(-) and SQuArE(+) refer to an AVA model only using negative references and a SQuArE model only using positive references. \# Refs is the total number of references used for the metric. [1,5] refers to the number of references being randomly sampled ${\in}[1,5]$.}
\label{tab:ablations}
\end{table}

%% file: main.bbl
\begin{thebibliography}{36}
\expandafter\ifx\csname natexlab\endcsname\relax\def\natexlab#1{#1}\fi

\bibitem[{Almazrouei et~al.(2023)Almazrouei, Alobeidli, Alshamsi, Cappelli, Cojocaru, Debbah, Goffinet, Heslow, Launay, Malartic, Noune, Pannier, and Penedo}]{falcon40b}
Ebtesam Almazrouei, Hamza Alobeidli, Abdulaziz Alshamsi, Alessandro Cappelli, Ruxandra Cojocaru, Merouane Debbah, Etienne Goffinet, Daniel Heslow, Julien Launay, Quentin Malartic, Badreddine Noune, Baptiste Pannier, and Guilherme Penedo. 2023.
\newblock {Falcon-40B}: an open large language model with state-of-the-art performance.

\bibitem[{Bajaj et~al.(2018)Bajaj, Campos, Craswell, Deng, Gao, Liu, Majumder, McNamara, Mitra, Nguyen, Rosenberg, Song, Stoica, Tiwary, and Wang}]{bajaj2018ms}
Payal Bajaj, Daniel Campos, Nick Craswell, Li~Deng, Jianfeng Gao, Xiaodong Liu, Rangan Majumder, Andrew McNamara, Bhaskar Mitra, Tri Nguyen, Mir Rosenberg, Xia Song, Alina Stoica, Saurabh Tiwary, and Tong Wang. 2018.
\newblock \href {http://arxiv.org/abs/1611.09268} {Ms marco: A human generated machine reading comprehension dataset}.

\bibitem[{Bulian et~al.(2022)Bulian, Buck, Gajewski, Boerschinger, and Schuster}]{bulian2022tomayto}
Jannis Bulian, Christian Buck, Wojciech Gajewski, Benjamin Boerschinger, and Tal Schuster. 2022.
\newblock \href {https://doi.org/10.48550/ARXIV.2202.07654} {Tomayto, tomahto. beyond token-level answer equivalence for question answering evaluation}.
\newblock In \emph{Proceedings of the 2022 Conference on Empirical Methods in Natural Language Processing}, Abu Dhabi, United Arab Emirates. Association for Computational Linguistics.

\bibitem[{Cao et~al.(2020)Cao, Dong, Wu, and Cheung}]{cao_factual_2020}
Meng Cao, Yue Dong, Jiapeng Wu, and Jackie Chi~Kit Cheung. 2020.
\newblock \href {https://doi.org/10.18653/v1/2020.emnlp-main.506} {Factual error correction for abstractive summarization models}.
\newblock In \emph{Proceedings of the 2020 Conference on Empirical Methods in Natural Language Processing (EMNLP)}, pages 6251--6258, Online. Association for Computational Linguistics.

\bibitem[{Chen et~al.(2019)Chen, Stanovsky, Singh, and Gardner}]{chen-etal-2019-evaluating}
Anthony Chen, Gabriel Stanovsky, Sameer Singh, and Matt Gardner. 2019.
\newblock \href {https://doi.org/10.18653/v1/D19-5817} {Evaluating question answering evaluation}.
\newblock In \emph{Proceedings of the 2nd Workshop on Machine Reading for Question Answering}, pages 119--124, Hong Kong, China. Association for Computational Linguistics.

\bibitem[{Clark et~al.(2019)Clark, Celikyilmaz, and Smith}]{clark-etal-2019-sentence}
Elizabeth Clark, Asli Celikyilmaz, and Noah~A. Smith. 2019.
\newblock \href {https://doi.org/10.18653/v1/P19-1264} {Sentence mover{'}s similarity: Automatic evaluation for multi-sentence texts}.
\newblock In \emph{Proceedings of the 57th Annual Meeting of the Association for Computational Linguistics}, pages 2748--2760, Florence, Italy. Association for Computational Linguistics.

\bibitem[{Di~Liello et~al.(2023)Di~Liello, Garg, and Moschitti}]{di-liello-etal-2023-context}
Luca Di~Liello, Siddhant Garg, and Alessandro Moschitti. 2023.
\newblock \href {https://doi.org/10.18653/v1/2023.acl-short.40} {Context-aware transformer pre-training for answer sentence selection}.
\newblock In \emph{Proceedings of the 61st Annual Meeting of the Association for Computational Linguistics (Volume 2: Short Papers)}, pages 458--468, Toronto, Canada. Association for Computational Linguistics.

\bibitem[{Di~Liello et~al.(2022)Di~Liello, Garg, Soldaini, and Moschitti}]{di-liello-etal-2022-pre}
Luca Di~Liello, Siddhant Garg, Luca Soldaini, and Alessandro Moschitti. 2022.
\newblock \href {https://doi.org/10.18653/v1/2022.emnlp-main.810} {Pre-training transformer models with sentence-level objectives for answer sentence selection}.
\newblock In \emph{Proceedings of the 2022 Conference on Empirical Methods in Natural Language Processing}, pages 11806--11816, Abu Dhabi, United Arab Emirates. Association for Computational Linguistics.

\bibitem[{Gabburo et~al.(2023)Gabburo, Garg, Koncel-Kedziorski, and Moschitti}]{gabburo-etal-2023-learning}
Matteo Gabburo, Siddhant Garg, Rik Koncel-Kedziorski, and Alessandro Moschitti. 2023.
\newblock \href {https://doi.org/10.18653/v1/2023.acl-long.467} {Learning answer generation using supervision from automatic question answering evaluators}.
\newblock In \emph{Proceedings of the 61st Annual Meeting of the Association for Computational Linguistics (Volume 1: Long Papers)}, pages 8389--8403, Toronto, Canada. Association for Computational Linguistics.

\bibitem[{Gabburo et~al.(2022)Gabburo, Koncel-Kedziorski, Garg, Soldaini, and Moschitti}]{gabburo2022}
Matteo Gabburo, Rik Koncel-Kedziorski, Siddhant Garg, Luca Soldaini, and Alessandro Moschitti. 2022.
\newblock \href {https://doi.org/10.18653/v1/2022.emnlp-main.645} {Knowledge transfer from answer ranking to answer generation}.
\newblock In \emph{Proceedings of the 2022 Conference on Empirical Methods in Natural Language Processing}, pages 9481--9495, Abu Dhabi, United Arab Emirates. Association for Computational Linguistics.

\bibitem[{Garg et~al.(2020)Garg, Vu, and Moschitti}]{garg2020tanda}
Siddhant Garg, Thuy Vu, and Alessandro Moschitti. 2020.
\newblock \href {https://doi.org/10.1609/aaai.v34i05.6282} {Tanda: Transfer and adapt pre-trained transformer models for answer sentence selection}.
\newblock \emph{Proceedings of the AAAI Conference on Artificial Intelligence}, 34(05):7780--7788.

\bibitem[{He et~al.(2021)He, Gao, and Chen}]{he2021debertav3}
Pengcheng He, Jianfeng Gao, and Weizhu Chen. 2021.
\newblock \href {http://arxiv.org/abs/2111.09543} {Debertav3: Improving deberta using electra-style pre-training with gradient-disentangled embedding sharing}.

\bibitem[{Hsu et~al.(2021)Hsu, Lind, Soldaini, and Moschitti}]{hsu2021answer}
Chao-Chun Hsu, Eric Lind, Luca Soldaini, and Alessandro Moschitti. 2021.
\newblock \href {https://doi.org/10.18653/v1/2021.findings-acl.374} {Answer generation for retrieval-based question answering systems}.
\newblock In \emph{Findings of the Association for Computational Linguistics: ACL-IJCNLP 2021}, pages 4276--4282, Online. Association for Computational Linguistics.

\bibitem[{Kingma and Ba(2015)}]{Kingma2015AdamAM}
Diederik~P. Kingma and Jimmy Ba. 2015.
\newblock \href {http://dblp.uni-trier.de/db/conf/iclr/iclr2015.html#KingmaB14} {Adam: A method for stochastic optimization.}
\newblock In \emph{International Conference of Learning Representations}.

\bibitem[{Kusner et~al.(2015)Kusner, Sun, Kolkin, and Weinberger}]{kusner2015wmd}
Matt~J. Kusner, Yu~Sun, Nicholas~I. Kolkin, and Kilian~Q. Weinberger. 2015.
\newblock From word embeddings to document distances.
\newblock In \emph{Proceedings of the 32nd International Conference on International Conference on Machine Learning - Volume 37}, ICML'15, page 957–966. JMLR.org.

\bibitem[{Lhoest et~al.(2021)Lhoest, Villanova~del Moral, Jernite, Thakur, von Platen, Patil, Chaumond, Drame, Plu, Tunstall, Davison, {\v{S}}a{\v{s}}ko, Chhablani, Malik, Brandeis, Le~Scao, Sanh, Xu, Patry, McMillan-Major, Schmid, Gugger, Delangue, Matussi{\`e}re, Debut, Bekman, Cistac, Goehringer, Mustar, Lagunas, Rush, and Wolf}]{lhoest-etal-2021-datasets}
Quentin Lhoest, Albert Villanova~del Moral, Yacine Jernite, Abhishek Thakur, Patrick von Platen, Suraj Patil, Julien Chaumond, Mariama Drame, Julien Plu, Lewis Tunstall, Joe Davison, Mario {\v{S}}a{\v{s}}ko, Gunjan Chhablani, Bhavitvya Malik, Simon Brandeis, Teven Le~Scao, Victor Sanh, Canwen Xu, Nicolas Patry, Angelina McMillan-Major, Philipp Schmid, Sylvain Gugger, Cl{\'e}ment Delangue, Th{\'e}o Matussi{\`e}re, Lysandre Debut, Stas Bekman, Pierric Cistac, Thibault Goehringer, Victor Mustar, Fran{\c{c}}ois Lagunas, Alexander Rush, and Thomas Wolf. 2021.
\newblock \href {https://doi.org/10.18653/v1/2021.emnlp-demo.21} {Datasets: A community library for natural language processing}.
\newblock In \emph{Proceedings of the 2021 Conference on Empirical Methods in Natural Language Processing: System Demonstrations}, pages 175--184, Online and Punta Cana, Dominican Republic. Association for Computational Linguistics.

\bibitem[{Lin(2004)}]{lin-2004-rouge}
Chin-Yew Lin. 2004.
\newblock \href {https://aclanthology.org/W04-1013} {{ROUGE}: A package for automatic evaluation of summaries}.
\newblock In \emph{Text Summarization Branches Out}, pages 74--81, Barcelona, Spain. Association for Computational Linguistics.

\bibitem[{Min et~al.(2021)Min, Boyd-Graber, Alberti, Chen, Choi, Collins, Guu, Hajishirzi, Lee, Palomaki, Raffel, Roberts, Kwiatkowski, Lewis, Wu, K\"uttler, Liu, Minervini, Stenetorp, Riedel, Yang, Seo, Izacard, Petroni, Hosseini, Cao, Grave, Yamada, Shimaoka, Suzuki, Miyawaki, Sato, Takahashi, Suzuki, Fajcik, Docekal, Ondrej, Smrz, Cheng, Shen, Liu, He, Chen, Gao, Oguz, Chen, Karpukhin, Peshterliev, Okhonko, Schlichtkrull, Gupta, Mehdad, and Yih}]{pmlr-v133-min21a}
Sewon Min, Jordan Boyd-Graber, Chris Alberti, Danqi Chen, Eunsol Choi, Michael Collins, Kelvin Guu, Hannaneh Hajishirzi, Kenton Lee, Jennimaria Palomaki, Colin Raffel, Adam Roberts, Tom Kwiatkowski, Patrick Lewis, Yuxiang Wu, Heinrich K\"uttler, Linqing Liu, Pasquale Minervini, Pontus Stenetorp, Sebastian Riedel, Sohee Yang, Minjoon Seo, Gautier Izacard, Fabio Petroni, Lucas Hosseini, Nicola~De Cao, Edouard Grave, Ikuya Yamada, Sonse Shimaoka, Masatoshi Suzuki, Shumpei Miyawaki, Shun Sato, Ryo Takahashi, Jun Suzuki, Martin Fajcik, Martin Docekal, Karel Ondrej, Pavel Smrz, Hao Cheng, Yelong Shen, Xiaodong Liu, Pengcheng He, Weizhu Chen, Jianfeng Gao, Barlas Oguz, Xilun Chen, Vladimir Karpukhin, Stan Peshterliev, Dmytro Okhonko, Michael Schlichtkrull, Sonal Gupta, Yashar Mehdad, and Wen-tau Yih. 2021.
\newblock \href {https://proceedings.mlr.press/v133/min21a.html} {Neurips 2020 efficientqa competition: Systems, analyses and lessons learned}.
\newblock In \emph{Proceedings of the NeurIPS 2020 Competition and Demonstration Track}, volume 133 of \emph{Proceedings of Machine Learning Research}, pages 86--111. PMLR.

\bibitem[{Muller et~al.(2021)Muller, Soldaini, Koncel{-}Kedziorski, Lind, and Moschitti}]{DBLP:journals/corr/abs-2110-07150}
Benjamin Muller, Luca Soldaini, Rik Koncel{-}Kedziorski, Eric Lind, and Alessandro Moschitti. 2021.
\newblock \href {http://arxiv.org/abs/2110.07150} {Cross-lingual genqa: {A} language-agnostic generative question answering approach for open-domain question answering}.
\newblock \emph{CoRR}, abs/2110.07150.

\bibitem[{Papineni et~al.(2001)Papineni, Roukos, Ward, and Zhu}]{papineni_bleu_2001}
Kishore Papineni, Salim Roukos, Todd Ward, and Wei-Jing Zhu. 2001.
\newblock \href {https://doi.org/10.3115/1073083.1073135} {{BLEU}: a method for automatic evaluation of machine translation}.
\newblock In \emph{Proceedings of the 40th Annual Meeting on Association for Computational Linguistics - {ACL} '02}, page 311. Association for Computational Linguistics.

\bibitem[{Paszke et~al.(2019)Paszke, Gross, Massa, Lerer, Bradbury, Chanan, Killeen, Lin, Gimelshein, Antiga, Desmaison, Kopf, Yang, DeVito, Raison, Tejani, Chilamkurthy, Steiner, Fang, Bai, and Chintala}]{pytorch}
Adam Paszke, Sam Gross, Francisco Massa, Adam Lerer, James Bradbury, Gregory Chanan, Trevor Killeen, Zeming Lin, Natalia Gimelshein, Luca Antiga, Alban Desmaison, Andreas Kopf, Edward Yang, Zachary DeVito, Martin Raison, Alykhan Tejani, Sasank Chilamkurthy, Benoit Steiner, Lu~Fang, Junjie Bai, and Soumith Chintala. 2019.
\newblock \href {http://papers.neurips.cc/paper/9015-pytorch-an-imperative-style-high-performance-deep-learning-library.pdf} {Pytorch: An imperative style, high-performance deep learning library}.
\newblock In \emph{Advances in Neural Information Processing Systems 32}, pages 8024--8035. Curran Associates, Inc.

\bibitem[{Rei et~al.(2020)Rei, Stewart, Farinha, and Lavie}]{rei_comet_2020}
Ricardo Rei, Craig Stewart, Ana~C Farinha, and Alon Lavie. 2020.
\newblock \href {https://doi.org/10.18653/v1/2020.emnlp-main.213} {{COMET}: A neural framework for {MT} evaluation}.
\newblock In \emph{Proceedings of the 2020 Conference on Empirical Methods in Natural Language Processing ({EMNLP})}, pages 2685--2702. Association for Computational Linguistics.

\bibitem[{Reiter(2018)}]{reiter_structured_2018}
Ehud Reiter. 2018.
\newblock \href {https://doi.org/10.1162/coli_a_00322} {{A Structured Review of the Validity of BLEU}}.
\newblock \emph{Computational Linguistics}, 44(3):393--401.

\bibitem[{Sellam et~al.(2020)Sellam, Das, and Parikh}]{sellam_bleurt_2020}
Thibault Sellam, Dipanjan Das, and Ankur Parikh. 2020.
\newblock \href {https://doi.org/10.18653/v1/2020.acl-main.704} {{BLEURT}: Learning robust metrics for text generation}.
\newblock In \emph{Proceedings of the 58th Annual Meeting of the Association for Computational Linguistics}, pages 7881--7892. Association for Computational Linguistics.

\bibitem[{Si et~al.(2021)Si, Zhao, and Boyd-Graber}]{si-etal-2021-whats}
Chenglei Si, Chen Zhao, and Jordan Boyd-Graber. 2021.
\newblock \href {https://doi.org/10.18653/v1/2021.emnlp-main.757} {What{'}s in a name? answer equivalence for open-domain question answering}.
\newblock In \emph{Proceedings of the 2021 Conference on Empirical Methods in Natural Language Processing}, pages 9623--9629, Online and Punta Cana, Dominican Republic. Association for Computational Linguistics.

\bibitem[{Virtanen et~al.(2020)Virtanen, Gommers, Oliphant, Haberland, Reddy, Cournapeau, Burovski, Peterson, Weckesser, Bright, {van der Walt}, Brett, Wilson, Millman, Mayorov, Nelson, Jones, Kern, Larson, Carey, Polat, Feng, Moore, {VanderPlas}, Laxalde, Perktold, Cimrman, Henriksen, Quintero, Harris, Archibald, Ribeiro, Pedregosa, {van Mulbregt}, and {SciPy 1.0 Contributors}}]{scipy}
Pauli Virtanen, Ralf Gommers, Travis~E. Oliphant, Matt Haberland, Tyler Reddy, David Cournapeau, Evgeni Burovski, Pearu Peterson, Warren Weckesser, Jonathan Bright, St{\'e}fan~J. {van der Walt}, Matthew Brett, Joshua Wilson, K.~Jarrod Millman, Nikolay Mayorov, Andrew R.~J. Nelson, Eric Jones, Robert Kern, Eric Larson, C~J Carey, {\.I}lhan Polat, Yu~Feng, Eric~W. Moore, Jake {VanderPlas}, Denis Laxalde, Josef Perktold, Robert Cimrman, Ian Henriksen, E.~A. Quintero, Charles~R. Harris, Anne~M. Archibald, Ant{\^o}nio~H. Ribeiro, Fabian Pedregosa, Paul {van Mulbregt}, and {SciPy 1.0 Contributors}. 2020.
\newblock \href {https://doi.org/10.1038/s41592-019-0686-2} {{{SciPy} 1.0: Fundamental Algorithms for Scientific Computing in Python}}.
\newblock \emph{Nature Methods}, 17:261--272.

\bibitem[{Vu and Moschitti(2021)}]{vu_ava_2021}
Thuy Vu and Alessandro Moschitti. 2021.
\newblock \href {https://doi.org/10.18653/v1/2021.naacl-main.412} {{AVA}: an automatic {eValuation} approach for question answering systems}.
\newblock In \emph{Proceedings of the 2021 Conference of the North American Chapter of the Association for Computational Linguistics: Human Language Technologies}, pages 5223--5233. Association for Computational Linguistics.

\bibitem[{Wang et~al.(2007)Wang, Smith, and Mitamura}]{wang2007jeopardy}
Mengqiu Wang, Noah~A Smith, and Teruko Mitamura. 2007.
\newblock What is the jeopardy model? a quasi-synchronous grammar for qa.
\newblock In \emph{Proceedings of the 2007 Joint Conference on Empirical Methods in Natural Language Processing and Computational Natural Language Learning (EMNLP-CoNLL)}, pages 22--32.

\bibitem[{Wegmann and Nguyen(2021)}]{wegmann_does_2021}
Anna Wegmann and Dong Nguyen. 2021.
\newblock \href {https://doi.org/10.18653/v1/2021.emnlp-main.569} {Does it capture {STEL}? a modular, similarity-based linguistic style evaluation framework}.
\newblock In \emph{Proceedings of the 2021 Conference on Empirical Methods in Natural Language Processing}, pages 7109--7130, Online and Punta Cana, Dominican Republic. Association for Computational Linguistics.

\bibitem[{Wolf et~al.(2020)Wolf, Debut, Sanh, Chaumond, Delangue, Moi, Cistac, Rault, Louf, Funtowicz, Davison, Shleifer, von Platen, Ma, Jernite, Plu, Xu, Le~Scao, Gugger, Drame, Lhoest, and Rush}]{wolf-etal-2020-transformers}
Thomas Wolf, Lysandre Debut, Victor Sanh, Julien Chaumond, Clement Delangue, Anthony Moi, Pierric Cistac, Tim Rault, Remi Louf, Morgan Funtowicz, Joe Davison, Sam Shleifer, Patrick von Platen, Clara Ma, Yacine Jernite, Julien Plu, Canwen Xu, Teven Le~Scao, Sylvain Gugger, Mariama Drame, Quentin Lhoest, and Alexander Rush. 2020.
\newblock \href {https://doi.org/10.18653/v1/2020.emnlp-demos.6} {Transformers: State-of-the-art natural language processing}.
\newblock In \emph{Proceedings of the 2020 Conference on Empirical Methods in Natural Language Processing: System Demonstrations}, pages 38--45, Online. Association for Computational Linguistics.

\bibitem[{Yang et~al.(2018)Yang, Liu, Liu, Lyu, and Li}]{yang-etal-2018-adaptations}
An~Yang, Kai Liu, Jing Liu, Yajuan Lyu, and Sujian Li. 2018.
\newblock \href {https://doi.org/10.18653/v1/W18-2611} {Adaptations of {ROUGE} and {BLEU} to better evaluate machine reading comprehension task}.
\newblock In \emph{Proceedings of the Workshop on Machine Reading for Question Answering}, pages 98--104, Melbourne, Australia. Association for Computational Linguistics.

\bibitem[{Yang et~al.(2015)Yang, Yih, and Meek}]{yang2015wikiqa}
Yi~Yang, Wen-tau Yih, and Christopher Meek. 2015.
\newblock Wikiqa: A challenge dataset for open-domain question answering.
\newblock In \emph{Proceedings of the 2015 conference on empirical methods in natural language processing}, pages 2013--2018.

\bibitem[{Zeng et~al.(2021)Zeng, Chen, Xu, and Li}]{zeng_gradient-based_2021}
Zhiyuan Zeng, Jiaze Chen, Weiran Xu, and Lei Li. 2021.
\newblock \href {https://doi.org/10.18653/v1/2021.emnlp-main.337} {Gradient-based adversarial factual consistency evaluation for abstractive summarization}.
\newblock In \emph{Proceedings of the 2021 Conference on Empirical Methods in Natural Language Processing}, pages 4102--4108, Online and Punta Cana, Dominican Republic. Association for Computational Linguistics.

\bibitem[{Zhang et~al.(2020-02-24)Zhang, Kishore, Wu, Weinberger, and Artzi}]{zhang_bertscore_2020}
Tianyi Zhang, Varsha Kishore, Felix Wu, Kilian~Q. Weinberger, and Yoav Artzi. 2020-02-24.
\newblock \href {http://arxiv.org/abs/1904.09675 [cs]} {{BERTScore}: Evaluating text generation with {BERT}}.
\newblock Number: {arXiv}:1904.09675.

\bibitem[{Zhang et~al.(2021)Zhang, Vu, and Moschitti}]{zhang-etal-2021-joint}
Zeyu Zhang, Thuy Vu, and Alessandro Moschitti. 2021.
\newblock \href {https://doi.org/10.18653/v1/2021.acl-long.252} {Joint models for answer verification in question answering systems}.
\newblock In \emph{Proceedings of the 59th Annual Meeting of the Association for Computational Linguistics and the 11th International Joint Conference on Natural Language Processing (Volume 1: Long Papers)}, pages 3252--3262, Online. Association for Computational Linguistics.

\bibitem[{Zhang et~al.(2022)Zhang, Vu, and Moschitti}]{trecqazeyu}
Zeyu Zhang, Thuy Vu, and Alessandro Moschitti. 2022.
\newblock \href {http://arxiv.org/abs/2201.05981} {Double retrieval and ranking for accurate question answering}.
\newblock \emph{CoRR}, abs/2201.05981.

\end{thebibliography}
